\documentclass[runningheads]{llncs}
\usepackage{graphicx}
\usepackage{hyperref}

\usepackage{amssymb} 
\usepackage{amsmath} 
\usepackage{xcolor} 
\usepackage{tikz}
\usepackage{subfig}

\newcommand{\colWidth}{\ \ \ \ }
\newcommand{\firstColWidth}{\ \ \ \ \ \ \ \ }

\newcommand{\keyword}[1]{\textrm{\bfseries #1}}
\newcommand{\constant}[1]{\textrm{\sffamily #1}}
\newcommand{\dmreturn}{\keyword{return\ }}
\newcommand{\dmnil}{\constant{nil}}

\newcommand{\dmif}{\keyword{if\ }}
\newcommand{\dmto}{\keyword{\ to\ }}
\newcommand{\dmfor}{\keyword{for\ }}
\newcommand{\dmforeach}{\keyword{for\ each\ }}

\newcommand{\dmelse}{\keyword{else\ }}

\newcommand{\becomes}{\leftarrow}

\newcommand{\fun}[1]{\textsc{#1}}

\newcommand{\sorted}[1]{\mathbf{#1}}

\newcommand{\mtps}{\sorted{M}}

\newcommand{\alg}[1]{\fun{#1}}
\newcommand{\COSIATEC}{\alg{COSIATEC}}

\newcommand{\SIATECCompress}{\alg{SIATECCompress}}

\newcommand{\SIA}{\alg{SIA}}

\newcommand{\SIATEC}{\alg{SIATEC}}

\newcommand{\SIACT}{\alg{SIACT}}

\newcommand{\SIAR}{\alg{SIAR}}

\newcommand{\SIARCTCFP}{\alg{SIARCT-CFP}}

\newcommand{\RECURSIA}{\alg{RecurSIA}}

\newcommand{\RRT}{\alg{RRT}}

\newcommand{\SortFun}{\fun{Sort}}

\newcommand{\SortLexFun}{\SortFun_{\mathrm{Lex}}}

\newcommand{\concat}{\oplus}

\newcommand{\dataset}{D}

\newcommand{\coveredset}{\fun{COV}}
\newcommand{\GetCoveredSet}{\coveredset}

\newcommand{\pattern}{P}

\newcommand{\encoding}{\sorted{E}}
\newcommand{\coveredSet}{S}
\newcommand{\residualPointSet}{R}

\newcommand{\minSize}{s_{\min}}

\newcommand{\transformations}{\mathbf T}
\newcommand{\transformation}{f}

\newcommand{\transformationObjectBasisPairs}{\mathbf V}
\newcommand{\tc}{F}

\newcommand{\basisSize}{\beta}
\newcommand{\objectBases}{\mathbf B}
\newcommand{\perms}{\mathbf R}
\newcommand{\objIndex}{i}
\newcommand{\basis}{\mathbf{b}}
\newcommand{\objectBasis}{{\basis}_{\mathrm{obj}}}
\newcommand{\imageBasis}{{\basis}_{\mathrm{img}}}
\newcommand{\imgBasisPerm}{{\basis}_{\mathrm{img}}'}
\newcommand{\imgIndex}{j}
\newcommand{\perm}{\bf r}

\newcommand{\mtpIndex}{\mathbf{SM}}
\newcommand{\occurrenceSets}{\mathbf{OS}}
\newcommand{\occurrenceSet}{\mathbf{os}}
\newcommand{\mtpSizes}{\mathbf{s}}
\newcommand{\sortedOccurrenceSets}{\mathbf{SOS}}
\newcommand{\comparator}{\gamma}

\newcommand{\diffSet}{X}

\begin{document}
\title{Understanding and Compressing Music with Maximal Transformable Patterns
}
\titlerunning{Maximal Transformable Patterns in Music}
\author{David Meredith\orcidID{0000-0002-9601-5017}}
\institute{
Department of Architecture, Design and Media Technology,\\
Aalborg University, Rendsburggade 14, 9000 Aalborg, Denmark\\
\email{dave@create.aau.dk}\\
\url{http://personprofil.aau.dk/119171}\\
\url{http://www.titanmusic.com}}
\maketitle              
\begin{abstract}
\sloppy We present a polynomial-time algorithm that discovers all maximal patterns in a point set, $\dataset\subset\mathbb{R}^k$, that are related by transformations in a user-specified class, $\tc$, of bijections over $\mathbb{R}^k$. We also present a second algorithm that discovers the set of occurrences for each of these maximal patterns and then uses compact encodings of these occurrence sets to compute a losslessly compressed encoding of the input point set.
This encoding takes the form of a set of pairs, $E=\left\lbrace\left\langle P_1, T_1\right\rangle,\left\langle P_2, T_2\right\rangle,\ldots\left\langle P_{\ell}, T_{\ell}\right\rangle\right\rbrace$, where each $\langle P_i,T_i\rangle$ consists of a maximal pattern, $P_i\subseteq \dataset$, and a set, $T_i\subset\tc$, of transformations that map $P_i$ onto other subsets of $\dataset$. Each transformation is encoded by a vector of real values that uniquely identifies it within $\tc$ and the length of this vector is used as a measure of the complexity of $\tc$.
We evaluate the new compression algorithm with three transformation classes of differing complexity, on the task of classifying folk-song melodies into tune families. The most complex of the classes tested includes all combinations of the musical transformations of transposition, inversion, retrograde, augmentation and diminution. We found that broadening the transformation class improved performance on this task. However, it did not, on average, improve compression factor, which may be due to the datasets (in this case, folk-song melodies) being too short and simple to benefit from the potentially greater number of pattern relationships that are discoverable with larger transformation classes.

\end{abstract}

\section{Introduction}
\label{introduction-section}

\begin{figure}
	\centering
	\subfloat[Score]{\resizebox{.6\textwidth}{!}{\includegraphics{"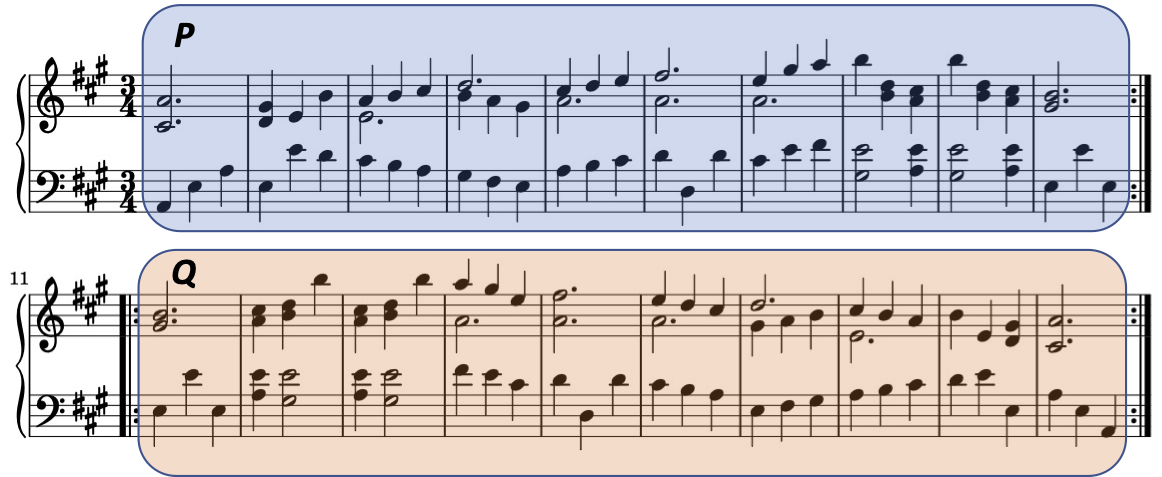"}}}\\
	\subfloat[Point set]{\resizebox{.7\textwidth}{!}{\includegraphics{"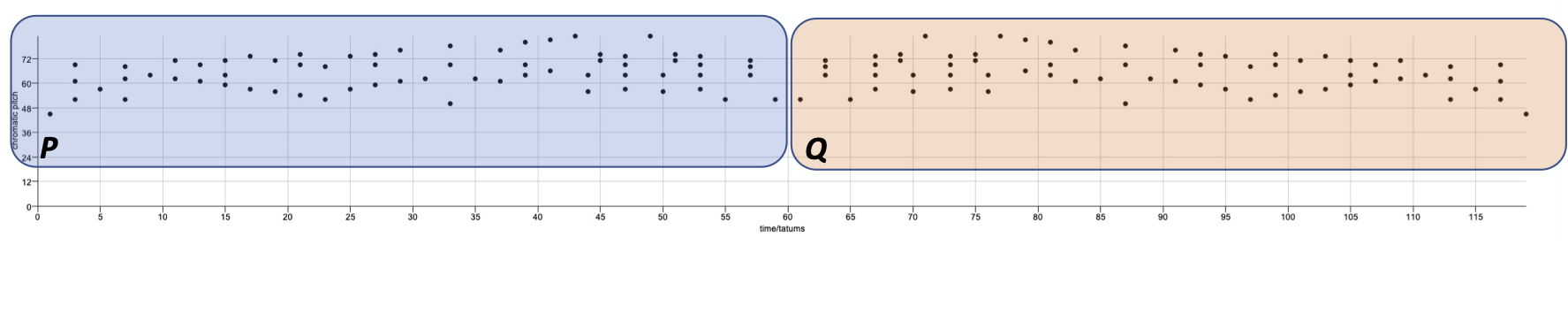"}}}
	\caption{\small Second movement of J.~Haydn's Sonata in A Major, Hob.XVI:26, Op.~13 No.~6 (``Menuetto al Rovescio''). The second half of the piece, $Q$, is an exact retrograde of the first half, $P$. 
	(b) shows a point set representing the score in (a), where each point represents a note, with the $x$-value indicating the temporal midpoint of the note and the $y$-value representing its chromatic pitch \cite[p.~127]{Meredith2006}.}
	\label{haydn}
\end{figure}


Suppose we want to understand a set of observations, $\dataset$, in the form of a set of $n$, $k$-dimensional points---that is, $\dataset\subset\mathbb{R}^k$. We call $\dataset$ a {\em dataset\/} and we want an explanation for $\dataset$, preferably in the form of 
an algorithm that outputs $\dataset$. There may be many different algorithms that output a given dataset. However, we expect only a small subset of these to faithfully describe the {\em actual\/} process that gave rise to the dataset. In science, to decide between two or more competing theories that account equally well for a set of observations, we often apply the parsimony principle and favour the simplest explanation, according to some measure of complexity.
We believe that, all else being equal, a simple explanation that accurately accounts for some observed data is less likely to do so by chance than an equally accurate, but more complex, explanation. 
We therefore want to find one of the {\em simplest\/} algorithms that outputs $\dataset$.  
We assume the dataset is given to us in the form of an {\em extensional\/} description in which each coordinate of each point is explicitly listed. Such a description represents the dataset as if it were a meaningless collection of unrelated observations. 
A description of a dataset never needs to be longer than its extensional description.

Suppose there is a set of points, $\pattern$, in our dataset, $\dataset$, that is mapped by a transformation, $f$, onto another set of points, $Q$, in $D$---that is, $Q=f(P)$. We call $\pattern$ and $Q$ {\em patterns\/} in $\dataset$.
We can now describe $P\cup Q$ by extensionally describing $P$ and then specifying $f$ instead of additionally extensionally describing $Q$. If $f$ can be described using fewer bits than are needed to extensionally describe $Q$, then a description in terms of $P$ and $f$ is a compressed encoding of $P\cup Q$. 
Moreover, by describing $P\cup Q$ as the result of applying the transformation $f$ to $P$, we make progress towards {\em explaining\/} $P\cup Q$, because we have now described how $Q$ might have arisen by transforming $P$. 

Now suppose that $\dataset\subset\mathbb{R}^2$ and that each point, $\langle t,p\rangle$, in $\dataset$ represents a note (or a sequence of tied notes) in a musical score, with $t$ representing the time at which the note occurs and $p$ representing its pitch. As an example, Fig.~\ref{haydn} shows  the ``Menuetto al Rovescio'' from Haydn's Sonata in A major, Hob.XVI:26. Note that pattern $Q$ in Fig.~\ref{haydn}(b) can be obtained by reflecting $P$ in the line $t=0$ (i.e., finding its retrograde) and then translating it parallel to the time axis. The full movement can be described by extensionally describing $P$ and then describing the transformation that maps $P$ onto $Q$. This gives a compressed description of the piece as well as an explanation for its second half.

\begin{figure}
	\centering
	\subfloat[Score]{\resizebox{.6\textwidth}{!}{\includegraphics{"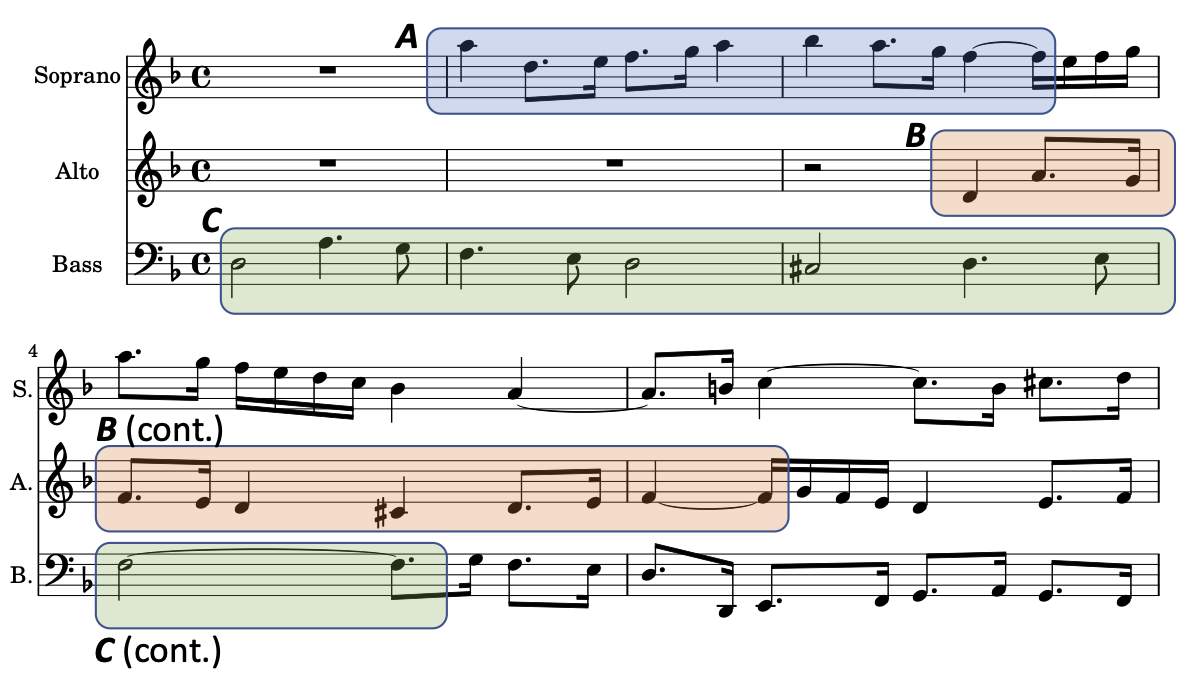"}}}\\
	\subfloat[Point set]{\resizebox{.7\textwidth}{!}{\includegraphics{"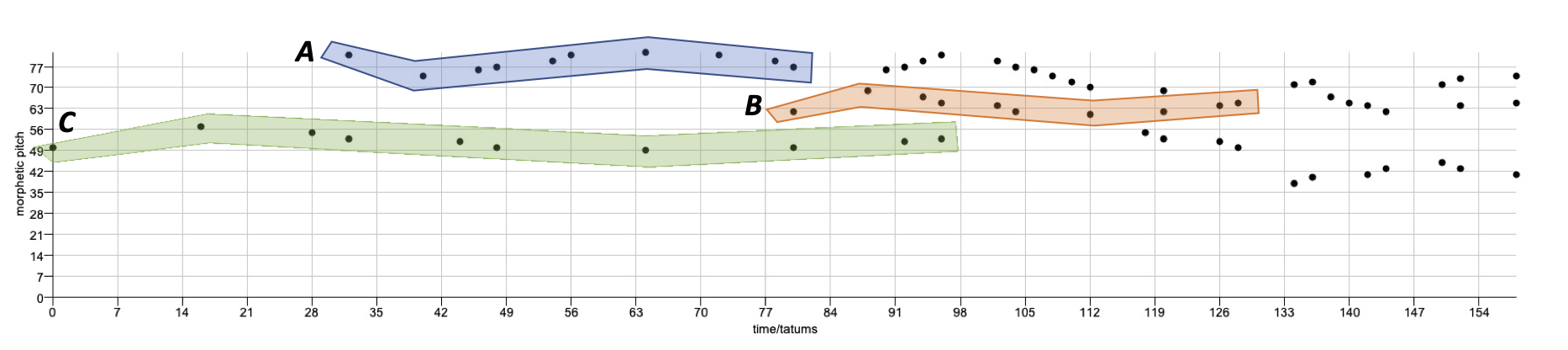"}}}
	\caption{\small Bars 1--5 of  {\em Contrapunctus VI\/} from J. S. Bach's {\em Die Kunst der Fuge}, BWV 1080. In (a) pattern $B$ is an inversion of pattern $A$ and pattern $C$ is an augmentation of pattern $B$. (b) shows a point set representing the passage, with each note or sequence of tied notes represented by a single point, $\langle t,p\rangle$, where $t$ represents the onset time of the note and $p$ is its morphetic pitch \cite[p.~127]{Meredith2006}. In (b), pattern $B$ can be obtained from pattern $A$ by a reflection in the time axis followed by a translation, while pattern $C$ can be obtained from pattern $B$ by stretching parallel to the time axis.}
	\label{contrapunctus}
\end{figure}

Figure~\ref{contrapunctus} shows another case where discovering simple geometric relationships between patterns is essential to understanding the music. In this extract from the opening of  {\em Contrapunctus VI\/} from J. S. Bach's {\em Die Kunst der Fuge}, BWV 1080, pattern $B$ is an inversion of pattern $A$, and pattern $C$ is an augmentation of pattern $B$. The set of points, $A\cup B\cup C$, can be described succinctly by extensionally defining $A$ and then specifying the transformations that map $A$ onto $B$ and $C$. 
This example illustrates that we can encode such a set of inter-related pattern occurrences as a pair, $\langle P,T\rangle$, where $P$ is an extensionally encoded pattern occurrence and $T$ is a set of transformations that map $P$ onto other occurrences of $P$ in the dataset. 
If each transformation in $T$ can be described using fewer bits than would be required to extensionally describe the occurrence that it generates from $P$, then we get a compressed encoding that we expect will represent a better way of understanding the data than that represented by an uncompressed, extensional description. 

Given a set of bijections, $F$, over $\mathbb{R}^k$, we say that a pattern, $P$, is {\em transformable\/} within a dataset, $D$, with respect to $F$, if there exists a transformation $f\in F$ such that $f(P)\subseteq D$. Because we believe in the parsimony principle, our goal is to find losslessly compressed encodings of datasets that are as short as possible, so we are primarily interested in finding {\em maximal\/} transformable patterns: we define a transformable pattern to be {\em maximal\/} within a dataset, $D$, for a transformation, $f\in F$, if there is no other pattern, $Q\subseteq D$, such that $f(Q)\subseteq D$ and $P\subset Q$.

In this paper, we present an algorithm that discovers the non-empty maximal transformable patterns (MTPs) in a dataset, $D\subset\mathbb{R}^k$, for transformations in a class, $F$, of bijections over $\mathbb{R}^k$. We also present an algorithm that computes a losslessly compressed encoding of $D$ in the form of a set of pairs, $\lbrace\langle P_1,T_1\rangle,\langle P_2,T_2\rangle,\ldots\langle P_{\ell},T_{\ell}\rangle\rbrace$, where each $P$ is an MTP in $D$ with respect to $F$, and $T$ is the set of transformations in $F$ that map $P$ onto subsets of $D$. 

In the next section, we review some related previous work. Then, in Section~\ref{theory-section}, we present some concepts and theory needed to understand the new algorithms, which will be described in Section~\ref{algorithms-section}. These algorithms could be used in a wide range of musical applications, including, for example, thematic and formal analysis, algorithmic composition, genre classification and composer recognition.  In Section~\ref{evaluation-section}, we present the results of an experiment in which we use the algorithms to automatically classify folk songs into tune families. Finally, in Section~\ref{conclusion-section}, we summarise the main contributions of this paper and suggest possible directions for future research.

\section{Related work}
\label{related-work-section}

Meredith, Lemstr{\"o}m and Wiggins \cite{MeredithLemstromWiggins2002}
defined a {\em maximal translatable pattern\/}  for a vector, $v$, in a point set, $D$, to be the set of points in $D$ that are mapped to other points in $D$ when translated by $v$. 
They 
presented an algorithm, \SIA, that computes the maximal translatable patterns in a set of $n$, $k$-dimensional points in $O(kn^2\log n)$ worst-case time and $O(kn^2)$ space. 
They defined two patterns to be {\em translationally equivalent\/} if and only if one can be obtained from the other by translation alone. They defined two patterns to be in the same {\em translational equivalence class\/} (TEC) if and only if they are translationally equivalent. 
They also presented an algorithm, \SIATEC, that computes the TEC for each maximal translatable pattern in a set of $n$, $k$-dimensional points in $O(kn^3)$ time and $O(kn^2)$ space. 
In \cite{MeredithLemstromWiggins2003}, Meredith, Lemstr{\"o}m and Wiggins 
introduced an algorithm called \COSIATEC\ that iteratively uses \SIATEC\ to compute a compressed description of a point set in the form of a set of TECs that collectively cover the input dataset. \COSIATEC\ achieves compression by encoding each TEC as an ordered pair, $\langle P,V\rangle$, where $P$ is one occurrence of the TEC's pattern and $V$ is the set of vectors that map $P$ onto its other occurrences in the dataset. If $T$ is a TEC and $C$ is the union of the occurrences in $T$, then the compression factor achieved by encoding $T$ in the form $\langle P,V\rangle$, as just described, is approximately $|C|/(|P|+|V|)$. $C$ is the TEC's {\em covered set\/} and $|C|$ is the {\em coverage\/} of the TEC.

Collins et al.~\cite{CollinsEtAl2010Short} claimed that maximal translatable patterns commonly consist of a musically important pattern along with other isolated points. To solve this problem, they proposed a ``compactness trawling" technique that involves searching within each maximal translatable pattern for subsets that are larger than some specified minimum size and more compact than some minimum compactness threshold. They combined this technique with \SIA\ to produce a new variant of the algorithm called \SIACT. 
In an attempt to improve on the precision and running time of \SIA, Collins \cite{Collins2011} designed a modified version of \SIA\ called \SIAR. Whereas \SIA\ computes inter-point vectors from each point to every lexicographically later point, \SIAR\ only computes the inter-point vectors from each point to the points that are not more than $r$ steps lexicographically later than it in the dataset, where $r$ is a parameter of the algorithm. This is approximately equivalent to running \SIA\ with a moving window.
In \cite{CollinsEtAl2013Short}, Collins et al.~address what they call the {\em inexactness problem\/} and the {\em precision problem}. The {\em inexactness problem\/} is that the basic versions of \SIA, \SIATEC\ and \COSIATEC\ are limited to only discovering exact occurrences of patterns under translation. The {\em precision problem\/} is that \SIA\ and \SIATEC\ typically discover many patterns in a dataset of which usually only a small fraction are of interest. In \COSIATEC, this problem is addressed by using heuristics (such as compactness, compression ratio and coverage) to select only the ``best'' TEC returned on each run of \SIATEC. Collins et al.~\cite{CollinsEtAl2013Short} propose an algorithm for computing inexactly related patterns, called \SIARCTCFP, in which they combine \SIAR\ and \SIACT\ with a categorisation and fingerprinting (CFP) technique. 
Collins et al. \cite{CollinsEtAlCMA2016} used \SIARCTCFP\ to discover interesting instances of thematic and harmonic re-use in Beethoven's piano sonatas.

Forth \cite{Forth2012} 
presented an algorithm that, like \COSIATEC, generates an encoding in the form of a set of TECs whose occurrences collectively cover 
the input point set. However, unlike \COSIATEC, Forth's algorithm runs \SIATEC\ only once and 
the covered sets of the TECs in these encodings may intersect. Forth's algorithm 
assigns a ``structural salience" to each of these TECs and then attempts to construct a cover consisting of structurally highly salient TECs.

In \cite{MeredithMIREX2013}, Meredith presented a compression algorithm based on \SIATEC, called \SIATECCompress. Like \COSIATEC\ and Forth's algorithm, \SIATECCompress\ computes a compressed encoding of a point set in the form of a set of maximal translatable pattern TECs that collectively cover the point set. Like Forth's algorithm, but unlike \COSIATEC, \SIATECCompress\ runs \SIATEC\ only once and computes covers in which the TEC covered sets may intersect. Like \COSIATEC, \SIATECCompress\ prefers TECs that individually have better compression factors, larger covered sets and more compact patterns. Typically, the encodings generated by \SIATECCompress\ are not as compressed as those computed by \COSIATEC. On the other hand, \SIATECCompress\ is faster than \COSIATEC\ because it only runs \SIATEC\ once.
Meredith \cite{MeredithJNMR2015} compared \COSIATEC, \SIATECCompress\ and Forth's algorithm on three musicological tasks: classifying folk tunes into tune families, discovery of repeated themes and sections, and detection of fugal subjects and countersubjects.
He also evaluated the effect of using \SIAR\ and/or \SIACT\ with each of the three basic algorithms. On the folk-song classification task, \COSIATEC\ performed best, while variants of \SIATECCompress\ performed best on the other two tasks.
More recently, in \cite{MeredithMML2019short}, two techniques were presented for improving the compression factor achievable using any TEC cover algorithm (such as \COSIATEC, \SIATECCompress\ or Forth's algorithm). The first of these techniques, {\em removal of redundant translators\/} (\RRT), involves analysing each TEC and removing as many occurrences as possible without reducing the TEC's covered set. The second technique, embodied in an algorithm called \RECURSIA, is to recursively apply the selected TEC cover algorithm to the patterns in each TEC and to replace the extensional description of each pattern in each $\langle P,V\rangle$ pair with a compressed encoding in terms of maximal translatable pattern TECs discovered {\em inside\/} that pattern. 

\section{Patterns and transformations in point sets}
\label{theory-section}

In this section we define some concepts required in order to understand the algorithms presented in Section~\ref{algorithms-section}.
We assume we are provided with a point set, $D\subset\mathbb{R}^k$, that we call a {\em dataset}.
%
%
We use the term {\em pattern\/} to refer to any set, $P\subset\mathbb{R}^k$, usually (but not always) under consideration in the context of a specified dataset. 
In the context of a dataset, $D\subset\mathbb{R}^k$,
we define a {\em transformation\/} to be a bijection, $f:\mathbb{R}^k\rightarrow\mathbb{R}^k$.
If a transformation, $f$, maps a point $p$ onto a point $q$, then we can write $q=f(p)$.
Bijections have unique inverses, therefore
the inverse, $f^{-1}$, of a transformation, $f$, is defined as follows: $f^{-1}(p) = q \textrm{ if and only if } f(q) = p$ for all $p,q\in\mathbb{R}^k$.
If $S\subset\mathbb{R}^k$ and $f:\mathbb{R}^k\rightarrow\mathbb{R}^k$ is a transformation, then, for convenience, we define that $f(S)=\bigcup_{p\in S}\lbrace f(p)\rbrace$ and $f^{-1}(S)=\bigcup_{p\in S}\lbrace f^{-1}(p)\rbrace$.

\subsection{Transformation classes}
\label{transformation-classes-section}
Given a dataset, $D\subset\mathbb{R}^k$,
we define a {\em transformation class}, $\tc$, to be some specified 
set of transformations over $\mathbb{R}^k$. We define each transformation, $f\in F$, to have a distinct {\em parameter}, $\alpha_F(f)$, that uniquely identifies $f$ within $F$. A transformation class, $F$, therefore has an associated {\em parameter set},
$
A(F) = \bigcup_{f\in F} \lbrace \alpha_F(f)\rbrace
$.
We define a function, $\phi_F$, such that, for each $f\in F$, there exists an expression, $\phi_F(\alpha_F(f),p)$, which is equal to $f(p)$ for all $p\in\mathbb{R}^k$. We call $\phi_F$ the {\em transformation class function\/} for the class $F$. We can therefore define any transformation class $F$ as follows:
$$
F=\lbrace f: (\exists\alpha\in A(F): \forall p\in\mathbb{R}^k,f(p)=\phi_F(\alpha,p))\rbrace\,.
$$
A transformation class, $F$, is therefore determined by its parameter set, $A(F)$, and its transformation class function, $\phi_F$.

As an example, let $F_{\textrm{2T}}$ be the set of 2-dimensional translations. If $\dataset\subset\mathbb{R}^2$ represents a passage of music like the datasets in Figs.~\ref{haydn} and \ref{contrapunctus}, then $F_{\textrm{2T}}$ contains the transformations that map patterns onto exact (but possibly transposed) restatements of those patterns.
Each transformation, $f\in F_{\textrm{2T}}$, is a function, $f:\mathbb{R}^2\rightarrow\mathbb{R}^2$, that translates a point, $p\in\mathbb{R}^2$, by a  vector, $v\in\mathbb{R}^2$. The vector $v$ uniquely determines $f$ within the transformation class, $F_{\textrm{2T}}$. That is, for each $f\in F_{\textrm{2T}}$, $\alpha_{F_{\textrm{2T}}}(f)=v$. 
The set of transformation parameters for this transformation class, $A(F_{\textrm{2T}})$, is therefore equal to $\mathbb{R}^2$ and the transformation class function is
$$\phi_{F_{\textrm{2T}}}(\alpha_{F_{\textrm{2T}}}(f),p)=p+\alpha_{F_{\textrm{2T}}}(f)\,.
$$
$F_{\textrm{2T}}$ can therefore be defined to be the set
$$
\lbrace f: (\exists\alpha\in A(F_{\textrm{2T}}): \forall p\in\mathbb{R}^2,f(p)=\phi_{F_{\textrm{2T}}}(\alpha_{F_{\textrm{2T}}}(f),p))\rbrace\,,
$$
which reduces to
$$
\lbrace f: (\exists v\in \mathbb{R}^2: \forall p\in\mathbb{R}^2,f(p)=p+v)\rbrace\,.
$$

Given a definition of the transformation class function, $\phi_F$, for a transformation class, $F$, we can uniquely specify any transformation, $f\in F$, by giving the transformation parameter for $f$ within $F$, $\alpha_F(f)$. 
A transformation parameter typically consists of some combination of vectors, matrices, or numerical values. 
However, it will always be possible to modify a transformation class function so that the transformation parameter is represented as a vector of real values, that the modified class function first parses in order to construct the 
transformation parameter.
We call such a vector a {\em transformation parameter vector\/} and we denote 
the transformation parameter vector corresponding to $\alpha_F(f)$
by $\sigma(\alpha_F(f))$.
 For notational convenience, we define that $\sigma(\alpha_F(f))=\sigma_F(f)$. The vector, $\sigma_F(f)$, contains all and only the independent numerical values required to uniquely specify $f$ within $F$. 
We define the {\em transformation class complexity}, $K(F)$, of a transformation class, $F$, to be the number of independent values in each of its transformation parameter vectors. That is, $K(F) = |\sigma_F(f)|$ for any $f\in F$. We define the {\em modified transformation class function}, $\phi_F'$, to be the modified form of $\phi_F$, the transformation class function, that first constructs the transformation parameter, $\alpha_F(f)$, from the transformation parameter vector, $\sigma_F(f)$, before applying $\phi_F$. In other words, $\phi_F'(\sigma_F(f),p)$ is always equal to $\phi_F(\alpha_F(f),p)$.
As a simple example, each transformation parameter for the class $F_{\textrm{2T}}$ is a 2-dimensional vector, which is already simply a vector of two independent real values. So for each function, $f$, in $F_{\textrm{2T}}$, we can define that
$\sigma_{F_{\textrm{2T}}}(f)=\alpha_{F_{\textrm{2T}}}(f)$ and therefore $\phi_{F_{\textrm{2T}}}$ is the same as $\phi_{F_{\textrm{2T}}}'$ and $K(F_{\textrm{2T}}) = 2$.

\subsection{Transformable patterns, occurrences and occurrence sets}
\label{transformable-patterns-section}
Given a dataset, $D\subset\mathbb{R}^k$, we define a pattern, $P\subseteq D$, to be {\em transformable\/} in $D$ with respect to a transformation class, $F$, if and only if there exists a transformation, $f\in F$, such that $f(P)\subseteq D$. If this is the case, we say that $P$ is {\em transformable by\/} $f$ in $D$ and that $f(P)$ is an {\em occurrence of\/} $P$ in $D$ with respect to $F$. We define the {\em occurrence set\/} of $P$ in $D$ with respect to $F$ to be the set,
$$\mathit{OS}(P,D,F)=\lbrace P\rbrace\cup\left\lbrace f(P): f(P)\subseteq D \textrm{ and } f\in F\right\rbrace\,.$$
Note that this definition of the term, ``occurrence set'', is more general than that which has been used in the MIREX Competitions on Discovery of Repeated Themes and Sections \cite{MIREXDiscoveryOfRepeatedThemes}. In these competitions, an ``occurrence set''  contains only those occurrences of a pattern that are inter-related by translation within pitch--time space. 
\sloppypar Given an occurrence set, $\mathcal{S}=\mathit{OS}(P,D,F)$, we define the {\em covered set\/} of $\mathcal{S}$ to be the set
$
\mathit{COV}(\mathcal{S})=\bigcup_{P\in\mathcal{S}} P
$.
An occurrence set, $\mathit{OS}(P,D,F)$, can be precisely specified by providing an extensional description of $P$ along with the transformation parameter of each transformation, $f\in F$, for which $f(P)\subseteq D$. This implies that, given $F$, we can describe $\mathit{COV}(\mathit{OS}(P,D,F))$ using at most $k|P|+K(F)(|\mathit{OS}(P,D,F)|-1)$ independent values. This compares with the $k|\mathit{COV}(\mathit{OS}(P,D,F))|$ independent values required to describe $\mathit{COV}(\mathit{OS}(P,D,F))$ extensionally. 
Disregarding the constant length of a given definition of $\phi'_F$, the {\em compression factor\/} that we can achieve with an occurrence set, $\mathit{OS}(P,D,F)$, is therefore at least
$$
\frac{k|\mathit{COV}(\mathit{OS}(P,D,F))|}{k|P|+K(F)(|\mathit{OS}(P,D,F)|-1)}\,.
$$
This compression factor is achieved by encoding the covered set, $\mathit{COV}(\mathit{OS}(P,D,F))$, by the ordered pair,
$$
\langle P,\lbrace\sigma: \phi_F'(\sigma,P)\subseteq D\land \phi_F'(\sigma,P)\neq P\rbrace\rangle\,,
$$
where $\phi_F'(\sigma,P)$ is a shorthand notation for $\bigcup_{p\in P}\lbrace\phi_F'(\sigma,p)\rbrace$. If we define 
$$\mathit\Psi = \lbrace\sigma: \phi_F'(\sigma,P)\subseteq D\land \phi_F'(\sigma,P)\neq P\rbrace\,,$$
then, given an encoding, $\langle P,\mathit\Psi\rangle$, the covered set itself can be reconstructed by evaluating the expression $P\cup\bigcup_{\sigma\in\mathit\Psi}\phi_F'(\sigma,P)$.

As an example, let $F_{\textrm{2TR}}$ be those transformations that consist of a 2-dimensional translation, optionally followed by a reflection in the $x$-axis. If the dataset represents a score of a piece of music, where each point represents a note or sequence of tied notes, the $x$-axis represents onset time and the $y$-axis represents pitch, then $F_{\textrm{2TR}}$ contains the transformations that map patterns onto exact repetitions, possibly inverted and/or transposed (e.g., the transformation that maps pattern $A$ onto pattern $B$ in Fig.~\ref{contrapunctus}). Each transformation, $f$, in $F_{\textrm{2TR}}$ is uniquely identified by a parameter, $\alpha_{F_{\textrm{2TR}}}(f)$ which is an ordered pair, $\langle v, b\rangle$, where $v$ is a 2-dimensional translation vector and $b$, which must be either $1$ or $-1$, is a scale factor for a stretch parallel to the $y$-axis, so that $b=-1$ indicates a reflection in the $x$-axis and $b=1$ indicates no reflection. In this case, 
if $\alpha=\langle v,b\rangle$, then we can define that $\sigma(\alpha)=\langle \alpha[0][0],\alpha[0][1],\alpha[1],\rangle$, so that $K(F_{\textrm{2TR}})=3$.
We can therefore define the modified transformation class function for this transformation class as follows:
$\phi_{F_{\textrm{2TR}}}'(\sigma,p)=\langle p[0]+\sigma[0],\sigma[2](p[1]+\sigma[1])\rangle$.

As another example, consider the class of 2-dimensional transformations that consist of a scaling parallel to the $x$-axis, followed by a translation, optionally followed by a reflection in the $x$-axis. We denote this class of transformations by $F_{\textrm{2STR}}$. Let $D\subset\mathbb{R}^2$, such that each point, $\langle t,p\rangle\in\dataset$, represents a note or sequence of tied notes, with $t$ representing the temporal mid-point of each note (or sequence of tied notes), and $p$ representing its pitch. $F_{\textrm{2STR}}$ then corresponds to the class of traditional contrapuntal transformations, consisting of any combination of transposition, inversion, retrograde, augmentation and diminution. $F_{\textrm{2STR}}$ contains the transformations that are required to describe the relationships between the patterns identified in the examples in Figs.~\ref{haydn} and \ref{contrapunctus}. Each transformation, $f$, in $F_{\textrm{2STR}}$ is uniquely specified by a transformation parameter, $\alpha_{F_\textrm{2STR}}(f)=\langle s, v, b\rangle$, where $s$ is a non-zero real value representing a scale factor for a stretch parallel to the $x$-axis (positive or negative so as to include retrogrades), $v\in\mathbb{R}^2$ is a 2-dimensional vector and $b\in\lbrace-1,1\rbrace$ is a scale factor for a stretch parallel to the $y$-axis, so that $b=-1$ indicates a reflection in the $x$-axis and $b=1$ indicates no reflection. The transformation class function for $F_{\textrm{2STR}}$ is
$$
	\phi_{F_{\textrm{2STR}}}(\alpha,p)=\langle p[0]\alpha[0]+\alpha[1][0],\alpha[2](p[1]+\alpha[1][1])\rangle\,.
$$ 
If $\alpha = \alpha_{F_{\textrm{2STR}}}(f)$ is the transformation parameter of a transformation, $f$, in $F_{\textrm{2STR}}$, then we can define the corresponding transformation parameter vector to be
$$
\sigma(\alpha)=\langle\alpha[0],\alpha[1][0],\alpha[1][1],\alpha[2]\rangle\,,
$$ 
which implies that $K(F_{\textrm{2STR}})=4$ and we can define the modified transformation class function for $F_{\textrm{2STR}}$ as follows:
$$
\phi_{F_{\textrm{2STR}}}'(\sigma,p)=\langle p[0]\sigma[0]+\sigma[1],\sigma[3](p[1]+\sigma[2])\rangle\,.
$$ 

\subsection{Maximal transformable patterns}
\label{maximal-transformable-patterns-section}
Given a dataset, $D$, and a transformation, $f$, we define the {\em maximal transformable pattern\/} (MTP) in $D$ for the transformation, $f$, to be the set of points,
$M(D,f)=D\cap f^{-1}(D)$.
It follows directly from this definition that
$
	M(D,f^{-1})=f(M(D,f))
$.
Note that this definition requires that a transformation $f$ has a well-defined inverse and is therefore a bijection.
The MTP for the transformation $f$ in the dataset $D$, denoted by $M(D,f)$, is thus the set of points in $D$ that are mapped by $f$ onto points in $D$. 
We say that a transformation, $f$, {\em occurs\/} in a dataset, $D$, if and only if $M(D,f)$ is non-empty.

As explained in Section~\ref{introduction-section} above, we can make progress towards both compressing and understanding a dataset, $D$, by discovering pairs of patterns in the dataset related by transformations that belong to transformation classes whose complexities are low. Suppose that $P$ and $Q$ are patterns in a dataset, $D\subset\mathbb{R}^k$, related by a transformation, $f$, in a transformation class, $F$, so that $P=f(Q)$. 
This allows us to describe $P\cup Q$ using $k|P|+K(F)$ independent values, compared with the $k|P\cup Q|$ values required to describe $P\cup Q$ extensionally, implying that we can potentially achieve a compression factor of $\frac{k|P\cup Q|}{k|P|+K(F)}$. For a given $f$, we therefore want to maximize $|P|$ and minimize $|P\cap Q|$---hence our interest in finding the {\em maximal\/} transformable pattern for each transformation, $f$, that occurs in a dataset.

There is a one-to-one correspondence between a maximal transformable pattern and a partial symmetry. A geometrical object has a particular type of symmetry if there is a transformation of that type (e.g., a reflection or a rotation) that maps the object onto itself. An object or point set, $D\subset\mathbb{R}^k$, is symmetrical (or invariant) with respect to some class of transformations, $F$, if there is a transformation, $f\in F$, such that $f(D)=D$. In this case, $f$ defines a {\em total symmetry\/} on $D$. More generally, a transformation, $g\in F$, defines a {\em partial symmetry\/} on $D$ if $g(D)\cap D\neq\emptyset$. It follows from the definition of an MTP given above, that a transformation, $f$, defines a partial symmetry on a dataset, $D$, if and only if its MTP in $D$, $M(D,f)$, is non-empty (i.e., iff $f$ occurs in $D$). If $M(D,f)=D$, then $f$ defines a total symmetry on $D$.

\section{Algorithms for discovering maximal transformable patterns and computing compressed encodings}
\label{algorithms-section}

We now present an algorithm for computing maximal transformable patterns (MTPs), together with a second algorithm that uses maximal transformable patterns to compute a compact encoding of a point set. Our pseudocode follows the conventions used in a number of related studies, e.g. \cite{MeredithJNMR2015}. Briefly, block structure is indicated by indentation alone and the assignment operator is `$\becomes$'. We denote ordered sets (aka. sequences, lists, vectors, arrays) using angle brackets, $\langle\cdot\rangle$, and we use zero-based indexing, so that, if $A$ is an ordered set, then $A[0]$ and $A[|A|-1]$ denote the first and last elements in $A$, respectively.
If we have two ordered sets, $A=\langle a_1, a_2, \ldots a_k \rangle$ and $B=\langle b_1, b_2, \ldots b_\ell \rangle$, then the {\em concatenation of\/} $B$ {\em onto} $A$ is
$A\oplus B=\langle a_1, a_2, \ldots a_k, b_1, b_2, \ldots b_\ell\rangle$. 
If $\mathit\Phi=\langle A_1,A_2,\ldots A_n\rangle$, where each $A_i$ is an ordered set, then we define that
$
\bigoplus_{i=1}^{n} A_i = \bigoplus_{A\in\mathit\Phi}A=A_1\oplus A_2\oplus\ldots \oplus A_{n}
$.
%

\begin{figure}[t]
	\begin{center}
		\resizebox*{.5\linewidth}{!}{
			\begin{minipage}{\linewidth}
				\begin{tabbing}
					\input{ColumnDefinitions}
					$\fun{MaximalTransformablePatterns}(\dataset,\tc,\minSize)$\\
					1\>$\transformationObjectBasisPairs\becomes\langle\rangle$\\
					2\>$\basisSize\becomes\fun{GetBasisSize}(\tc)$\\
					3\>$\objectBases\becomes\fun{ComputeObjectBases}(\dataset,\basisSize)$\\
					4\>$\perms\becomes\fun{ComputePermutationIndexSequences}(\basisSize)$\\
					5\>$\dmfor\objIndex\becomes0\dmto|\objectBases|-1$\\
					6\>\>$\objectBasis\becomes\objectBases[\objIndex]$\\
					7\>\>$\dmfor\imgIndex\becomes\objIndex\dmto|\objectBases|-1$\\
					8\>\>\>$\imageBasis\becomes\objectBases[\imgIndex]$\\
					9\>\>\>$\dmforeach\perm\in\perms$\\
					10\>\>\>\>$\imgBasisPerm\becomes\bigoplus_{\ell=0}^{\basisSize-1}\langle\imageBasis[\perm[\ell]]\rangle$\\
					11\>\>\>\>$\transformations\becomes\fun{GetTransformations}(\tc,\objectBasis,\imgBasisPerm)$\\
					12\>\>\>\>$\dmforeach\transformation\in\transformations$\\
					13\>\>\>\>\>$\transformationObjectBasisPairs\becomes\transformationObjectBasisPairs\oplus
						\langle\langle\transformation,\objectBasis\rangle,\langle\transformation^{-1},\imageBasis\rangle\rangle$\\
					14\>$\transformationObjectBasisPairs\becomes\SortLexFun(\transformationObjectBasisPairs)$\\
					15\>$\mtps\becomes\langle\rangle$\\
					16\>$\dmif|\transformationObjectBasisPairs|>0$\\
					17\>\>$\pattern\becomes\dmnil,\transformation\becomes\dmnil$\\
					18\>\>$\dmfor i\becomes0\dmto|\transformationObjectBasisPairs|-1$\\
					19\>\>\>$\dmif \pattern=\dmnil\land\transformation=\dmnil$\\
					20\>\>\>\>$\langle\transformation,\pattern\rangle\becomes\transformationObjectBasisPairs[i]$\\
					21\>\>\>$\dmelse\dmif\transformationObjectBasisPairs[i][0]=\transformation$\\
					22\>\>\>\>$\pattern\becomes\pattern\cup\transformationObjectBasisPairs[i][1]$\\
					23\>\>\>$\dmelse$\\
					24\>\>\>\>$\dmif|P|\ge\minSize$\\
					25\>\>\>\>\>$\mtps\becomes\mtps\concat\langle\langle\transformation,\pattern\rangle\rangle$\\
					26\>\>\>\>$\langle\transformation,\pattern\rangle\becomes\transformationObjectBasisPairs[i]$\\
					27\>\>$\dmif|P|\ge\minSize$\\
					28\>\>\>$\mtps\becomes\mtps\concat\langle\langle\transformation,\pattern\rangle\rangle$\\
					29\>$\dmreturn\mtps$\\
				\end{tabbing}
			\end{minipage}
		}
	\end{center}
	\caption{Algorithm for computing the maximal transformable patterns of size at least $\minSize$, in a dataset, $D$, with respect to a transformation class, $\tc$.}
	\label{compute-max-tran-pats-algorithm}
\end{figure}

\subsection{Computing the maximal transformable patterns in a point set}
\label{computing-mtps-section}

Figure~\ref{compute-max-tran-pats-algorithm} shows an algorithm that computes the non-empty maximal transformable patterns of size at least $\minSize$, in a dataset, $\dataset\subset\mathbb{R}^k$, with respect to a transformation class, $\tc$. When $\tc$ is a class of translations, the algorithm reduces to being \SIA\ \cite{MeredithLemstromWiggins2002}. The algorithm generalizes the strategy adopted in \SIA\ to the discovery of MTPs that are related by {\em any user-specified class of bijections over\/} $\mathbb{R}^k$.

Suppose we have two ordered sets of points, $\sorted P$ and $\sorted Q$, such that $|\sorted P| = |\sorted Q| = \basisSize$ and there exists at least one transformation, $f$, in a transformation class, $F$, such that $\sorted Q[i]=f(\sorted P[i])$ for all $0\le i<|\sorted P|$. For a given $\tc$, we can choose $\basisSize$ to be just large enough so that there is a manageably small maximum number of distinct transformations, $f\in F$, for which this is true. For example, for the transformation class, $F_{\mathrm{2T}}$, we can set $\basisSize$ to 1, since there is only one translation vector that maps any given single point onto any other single point. 
On the other hand, given two distinct points, $p$ and $q$, there are exactly {\em two\/} transformations, $g$, in $F_{\mathrm{2TR}}$, such that $q=g(p)$, namely those whose parameter vectors are $\langle q[0]-p[0],q[1]-p[1],1\rangle$ and $\langle q[0]-p[0],-p[1]-q[1],-1\rangle$. For the transformation class $F_{\mathrm{2STR}}$, if we set $\basisSize$ to 2, then there are at most 2 transformations that map $\sorted P$ to $\sorted Q$. We use the term {\em object basis\/} and {\em image basis\/} to refer to such point sequences, $\mathbf P$ and $\mathbf Q$, that are just large enough to determine a manageably small number of functions within some transformation class that map $\mathbf P$ to $\mathbf Q$. 

The algorithm in Fig.~\ref{compute-max-tran-pats-algorithm} first initializes the variable, $\transformationObjectBasisPairs$, which will hold a list of pairs, $\langle \transformation,\basis\rangle$, in which $\transformation$ is a transformation that belongs to the class, $\tc$, and $\basis$ is a basis that is transformable within $\dataset$ by $\transformation$.
In line 2 of the algorithm, we determine the basis size, $\basisSize$, for $\tc$.
Then, in line 3, the complete set of $\frac{|\dataset|!}{{\basisSize!(|\dataset|-\basisSize)!}}=O(\frac{|\dataset|^{\basisSize}}{\basisSize!})$ $\basisSize$--combinations in $\dataset$ is computed and each combination is represented as an {\em ordered\/} set in which the points are in lexicographical order. Each of these ordered sets of points is an object basis and the complete set of object bases is stored in $\objectBases$.
In line 4, the set of $\basisSize!$ permutations of the sequence $\langle0,\ldots\basisSize-1\rangle$ is computed and stored in lexicographical order in $\perms$. The basis size, $\basisSize$, will typically be small (in the cases considered in this paper, $\basisSize\le 2$), so $\basisSize!$ is also typically small.
In lines 5--11, we take each object basis, $\objectBasis\in\objectBases$ and then, for each permutation, $\imgBasisPerm$, of each basis, $\imageBasis$, that does not occur before $\objectBasis$ in $\objectBases$, we compute (in line 11) the transformations in $\tc$ that map $\objectBasis$ onto $\imgBasisPerm$. 
In lines 12--13, for each of these computed transformations, $\transformation\in\transformations$, we append both $\langle\transformation,\objectBasis\rangle$ and $\langle\transformation^{-1},\imageBasis\rangle$ to $\transformationObjectBasisPairs$.
In line 14, $\transformationObjectBasisPairs$ is sorted lexicographically, so that $\langle\mathrm{transformation},\mathrm{basis}\rangle$ pairs in $\transformationObjectBasisPairs$ with the same transformation form contiguous segments.
The MTP for each transformation, $\transformation\in\tc$, that occurs in $\dataset$ is then equal to the union of the bases in the contiguous pairs in $\transformationObjectBasisPairs$ whose first elements are equal to $\transformation$. 
The non-empty MTPs can therefore be found by scanning $\transformationObjectBasisPairs$ just once using the process described in lines 17--28. For a $k$--dimensional dataset of size $n$, the overall running time is $O\left(\frac{k\tau}{(\basisSize-1)!} n^{2\basisSize}\log n\right)$ where $\tau$ is the maximum number of transformations returned by the $\fun{GetTransformations}$ function called in line 11. 
The algorithm uses $O\left(\frac{\tau(K(F)+k\basisSize)}{\basisSize!} n^{2\basisSize}\right)$ space.

\subsection{Compressing point sets using maximal transformable patterns}
\label{compressing-point-sets-section}

Figure~\ref{encode-point-set} outlines an algorithm that computes a compressed encoding of a point set in the form of a set of MTP occurrence sets. It takes three arguments: the input dataset, $\dataset$, a transformation class, $\tc$, and a minimum size, $\minSize$, for the MTPs used in the encoding. The algorithm uses a simple greedy strategy, similar to that used in Forth's algorithm \cite{Forth2012} and SIATECCompress \cite{MeredithMIREX2013}, to find a set of MTP occurrence sets with high compression factors, whose covered sets collectively cover the input dataset. 
The first step is to use the algorithm in Fig.~\ref{compute-max-tran-pats-algorithm} to compute the non-empty MTPs in $\dataset$ for the transformations in $\tc$ (line 1 in Fig.~\ref{encode-point-set}). The MTPs are stored in a list, $\mtps$, in which each element is an ordered pair, $\langle f,P\rangle$, where $P$ is the MTP in $\dataset$ for the transformation, $f$. 
Next, in line 2, the MTPs are indexed by their size in a data structure, $\mtpIndex$, which is an array of $|\dataset|+1$ lists. At the conclusion of line 2, each $\mtpIndex[i]$ is a list of the MTPs whose size is $i$, sorted lexicographically by their patterns, so that the MTPs with a given pattern, $\pattern$, form a contiguous segment. The variable $\mtpSizes$ contains the set of distinct sizes of MTPs discovered, in increasing order. The set of MTPs of a given size can therefore be accessed in constant time and we can use $\mtpSizes$ to iterate over just the non-empty elements in $\mtpIndex$ in $O(|\mtpSizes|)$ time.  

\begin{figure}[t]
	\begin{center}
		\resizebox*{.6\linewidth}{!}{
			\begin{minipage}{\linewidth}
				\begin{tabbing}
					\input{ColumnDefinitions}
					$\fun{EncodePointSet}(\dataset,\tc,\minSize)$\\
					1\>$\mtps\becomes\fun{MaximalTransformablePatterns}(\dataset,\tc,\minSize)$\\
					2\>$\langle\mtpIndex,\mtpSizes\rangle\becomes\fun{IndexMTPs}(\mtps,|\dataset|)$\\
					3\>$\occurrenceSets\becomes\fun{MergeMTPs}(\mtpIndex,\mtpSizes,|\dataset|)$\\
					4\>$\occurrenceSets\becomes\fun{ComputeOccurrenceSets}(\occurrenceSets,\mtpIndex,\mtpSizes)$\\
					5\>$\sortedOccurrenceSets\becomes\fun{SortOccurrenceSets}(\occurrenceSets,\comparator,\mtpSizes)$\\
					6\>$\dmreturn\fun{ComputeEncoding}(\sortedOccurrenceSets)$
				\end{tabbing}
			\end{minipage}
		}
	\end{center}
	\caption{The $\fun{EncodePointSet}$ algorithm.}
	\label{encode-point-set}
\end{figure}

A single pattern can be the MTP for two or more transformations in a dataset. All the $\langle f,P\rangle$ pairs with a given pattern form a contiguous segment within one of the lists in $\mtpIndex$, 
so we can efficiently convert 
each such cluster
into a pair, $\langle P, T\rangle$, where $T$ is the set of transformations for which $P$ is an MTP in $\dataset$. 
This is done in line 3 
using the $\fun{MergeMTPs}$ function, which constructs an array, $\occurrenceSets$, of $|\dataset|+1$ lists, that indexes the MTP occurrence sets in a similar way to that in which $\mtpIndex$ indexes the MTPs themselves. For each non-empty list of MTPs in $\mtpIndex$, $\fun{MergeMTPs}$ merges each cluster of same-pattern MTPs into a pair, $\occurrenceSet=\langle P,T\rangle$, that is added to the appropriate list in $\occurrenceSets$. 

If a pattern, $Q$, is transformable within $D$ by a transformation, $f$, then any subset of $Q$ will also be transformable within $D$ by $f$. Moreover, for every transformation, $f\in\tc$ that occurs in $\dataset$, there exists exactly one $\langle P, T\rangle$ pair in $\occurrenceSets$ whose pattern, $P$, is the MTP of $f$. The occurrence set within $D$ with respect to $\tc$ of an MTP, $P$, is therefore the pair,
$$\langle P,\lbrace f\mid(\exists\,\langle Q,T\rangle\in\occurrenceSets
\mid (f\in T\land P\subseteq Q))\rbrace\rangle\,,$$
where, to promote readability, we define $\langle Q,T\rangle\in\occurrenceSets$ if and only if there exists an $i$ such that $\langle Q,T\rangle\in\occurrenceSets[i]$. This implies that we can find the occurrence set of a pattern, $P$, simply by finding the union of the transformation sets of the pairs in $\occurrenceSets$ whose patterns contain $P$. This is the strategy adopted in the $\fun{ComputeOccurrenceSets}$ function shown in Fig.~\ref{compute-occurrence-sets} and called in line 4 of \fun{EncodePointSet}.

\begin{figure}
	\begin{center}
		\resizebox*{.6\linewidth}{!}{
			\begin{minipage}{\linewidth}
				\begin{tabbing}
					\input{ColumnDefinitions}
					$\fun{ComputeOccurrenceSets}(\occurrenceSets,\mtpIndex,\mtpSizes)$\\
					1\>$\dmfor i\becomes 1\dmto|\mtpSizes|-2$\\
					2\>\>$m\becomes\mtpSizes[i]$\\
					3\>\>$\dmforeach\occurrenceSet\in\occurrenceSets[m]$\\
					4\>\>\>$\dmfor j\becomes i+1\dmto|\mtpSizes|-1$\\
					5\>\>\>\>$\dmforeach\occurrenceSet_2\in\occurrenceSets[\mtpSizes[j]]$\\
					6\>\>\>\>\>$\dmif\occurrenceSet_2[0]\supset\occurrenceSet[0]$\\
					7\>\>\>\>\>\>$\occurrenceSet[1]\becomes\occurrenceSet[1]\cup\occurrenceSet_2[1]$\\
					8\>$\dmforeach m\in\mtpSizes$\\
					9\>\>$\fun{DeDupeAndSort}(\occurrenceSets[m])$\\
					10\>\>$\dmforeach\occurrenceSet\in\occurrenceSets[m]$\\
					11\>\>\>$\occurrenceSet\becomes\fun{RemoveRedundantTransformations}(\occurrenceSet)$\\
					12\>\>$\fun{RemoveOccurrenceSetsWithNoTransformations}(\occurrenceSets[m])$\\
					15\>$\dmreturn \occurrenceSets$\\
				\end{tabbing}
			\end{minipage}
		}
	\end{center}
	\caption{The $\fun{ComputeOccurrenceSets}$ function.}
	\label{compute-occurrence-sets}
\end{figure}

For each occurrence set, $\langle P, T\rangle$, in $\occurrenceSets$, $\fun{ComputeOccurrenceSets}$ first searches through the occurrence sets with patterns larger than $P$ for ones whose patterns contain $P$ (lines 1--7 in Fig.~\ref{compute-occurrence-sets}). When it finds such an occurrence set, $\langle Q,T'\rangle$, it sets $T$ to $T\cup T'$
(lines 6--7 in Fig.~\ref{compute-occurrence-sets}). 
This can result in occurrence sets that are equal to each other, even if their ordered-pair representations, $\langle P,T\rangle$, are different. In lines 8--9, the algorithm therefore iterates over the set of MTP sizes, $\mtpSizes$, and, for each size, $m$, it de-duplicates and sorts the list of occurrence sets, $\occurrenceSets[m]$. Redundant transformations are then removed from the transformation sets of the remaining occurrence sets in $\occurrenceSets$ (line 10--11). Given an occurrence set, $\langle P, T\rangle$, if there are two transformations in $T$ that map the object pattern, $P$, onto the same image pattern, then the more complex transformation is removed. 
For each occurrence set,  $\langle P, T\rangle$, the algorithm also uses an adaptation of the RRT algorithm described in \cite{MeredithMML2019short} to remove as many elements from $T$ as possible without reducing the covered set of $\langle P,T\rangle$. 
Finally, in line 12,
occurrence sets with empty transformation sets are removed, as they cannot contribute to compression.

When $\fun{ComputeOccurrenceSets}$ has finished executing in line 4 of \fun{EncodePointSet}, $\occurrenceSets$ contains MTP occurrence sets with compression factors greater than 1. The final step is to find a subset of these occurrence sets that collectively cover the dataset and whose combined description length is as low as we can make it using the simple, greedy compression strategy that we adopt here. 
In this strategy, we first sort the occurrence sets in $\occurrenceSets$ to produce a sorted list, $\sortedOccurrenceSets$, so that more preferred occurrence sets appear earlier in the list (line 5 of Fig.~\ref{encode-point-set}). In the implementation used to obtain the results reported below (Section~\ref{evaluation-section}), the occurrence sets were sorted into decreasing order by compression factor and then coverage. 
The \fun{ComputeEncoding} function, shown in Fig.~\ref{compute-encoding}, is then called  in line 6 of $\fun{EncodePointSet}$ to compute the final encoding of the dataset. \begin{figure}[t]
	\begin{center}
		\resizebox*{.4\linewidth}{!}{
			\begin{minipage}{\linewidth}
				\begin{tabbing}
					\input{ColumnDefinitions}
					$\fun{ComputeEncoding}(\sortedOccurrenceSets,\dataset)$\\
					1\>$\encoding\becomes\langle\sortedOccurrenceSets[0]\rangle$\\
					2\>$\coveredSet\becomes\GetCoveredSet(\sortedOccurrenceSets[0])$\\
					3\>$\dmfor i\becomes 1\dmto|\sortedOccurrenceSets|-1$\\
					4\>\>$\occurrenceSet\becomes\sortedOccurrenceSets[i]$\\
					5\>\>$\ell\becomes\fun{Length}(\occurrenceSet[0])+\fun{Length}(\occurrenceSet[1])$\\
					6\>\>$\diffSet\becomes\GetCoveredSet(\occurrenceSet)\setminus\coveredSet$\\
					7\>\>$\dmif\ell < \fun{Dim}(\dataset)\times|\diffSet|$\\
					8\>\>\>$\encoding\becomes\encoding\concat\langle\occurrenceSet\rangle$\\
					9\>\>\>$\coveredSet\becomes\coveredSet\cup\diffSet$\\
					10\>$\residualPointSet\becomes\dataset\setminus\coveredSet$\\
					11\>$\dmif\residualPointSet\neq\varnothing$\\
					12\>\>$\encoding\becomes\encoding\concat\langle\langle\residualPointSet,\varnothing\rangle\rangle$\\
					13\>$\dmreturn \encoding$
				\end{tabbing}
			\end{minipage}
		}
	\end{center}
	\caption{The $\fun{ComputeEncoding}$ function.}
	\label{compute-encoding}
\end{figure}

$\fun{ComputeEncoding}$ takes the sorted list of occurrence sets, $\sortedOccurrenceSets$, and the dataset, $\dataset$, and outputs an encoding in the form of a list, $\encoding$, of occurrence sets, each represented by a $\langle P,T\rangle$ pair, in which $P$ is an MTP and $T$ is a set of transformations in $\tc$ that map $P$ onto other occurrences of $P$ in $\dataset$.  The variable $\encoding$ is first initialized to hold the most preferred occurrence set, which is the first in $\sortedOccurrenceSets$ (line 1). 
In line 2, we then initialize a set, $\coveredSet$, which holds the accumulated set of points collectively covered by the occurrence sets added to $\encoding$. The algorithm iterates over the remaining occurrence sets in $\sortedOccurrenceSets$ (lines 3--9). The description length, $\ell$, of each $\occurrenceSet$ is computed (line 5) in terms of the total number of real-valued components in the points in its pattern and the parameter vectors in its transformation set. The set, $\diffSet$, is then computed, containing points in the covered set of $\occurrenceSet$ that are not yet present in $\coveredSet$. $\diffSet$ is the set of points that would be added to $\coveredSet$ if $\occurrenceSet$ were added to $\encoding$. If $\ell$ is less than the product of the dimension of $\dataset$ and the number of points in $\diffSet$, then $\occurrenceSet$ is added to $\encoding$ as this results in $\diffSet$ being encoded in a compressed form (line 7--8). The points in $\diffSet$ are then added to $\coveredSet$ (line 9). Finally, after iterating over all the occurrence sets in $\sortedOccurrenceSets$, there may remain some uncovered points, forming a so-called {\em residual point set\/} \cite{MeredithJNMR2015}. This can occur because $\sortedOccurrenceSets$ only contains occurrence sets whose compression factors are greater than one, as a result of non-compressing occurrence sets having been removed in line 12 of $\fun{ComputeOccurrenceSets}$ (see Fig.~\ref{compute-occurrence-sets}). In lines 10--12 of $\fun{ComputeEncoding}$, this residual point set is computed, represented as an occurrence set with an empty transformation set, and then appended to the encoding.

\section{Evaluation}
\label{evaluation-section}

$\fun{EncodePointSet}$ (Fig.~\ref{encode-point-set}) was evaluated on the task of classifying folk-song melodies into tune families, using the {\em Annotated Corpus} \cite{vanKranenburgVolkWiering2013} of 360 melodies from the Dutch folk song collection, {\em Onder de groene linde\/} \cite{Grijp2008}.
$\fun{EncodePointSet}$ was used as a compressor to calculate the normalized compression distance (NCD) \cite{LiEtAl2004} between each pair of melodies in the collection. Each melody was then classified using the 1-nearest-neighbour algorithm with leave-one-out cross-validation. The classifications obtained were compared with a ground-truth classification produced by expert musicologists. $\fun{EncodePointSet}$ was tested using the three transformation classes, $F_{\mathrm{2T}}$, $F_{\mathrm{2TR}}$ and $F_{\mathrm{2STR}}$, introduced above. 
We calculated the NCDs between the folk song melodies in the same way as that described in \cite{MeredithJNMR2015} and we evaluated performance in terms of average compression factor and classification success rate (see Table~\ref{results-table}).
\setlength{\tabcolsep}{6pt}
\renewcommand{\arraystretch}{1.1}
\begin{table}
\caption{\small Results of running $\fun{EncodePointSet}$ on the Annotated Corpus of the Dutch Folk Song Database with the three different transformation classes, $F_{\mathrm{2T}}$, $F_{\mathrm{2TR}}$ and $F_{\mathrm{2STR}}$. $\mathit{SR}$ is classification success rate. Columns 3 and 4 give the mean compression factor achieved over, respectively, the corpus and the file-pairs used to compute the compression distances.}
\label{results-table}
\vskip.1in
\centering
\begin{tabular}{c|ccc}
\hline
{\em Transformation class} & {\em SR} & {\em CF on corpus} & {\em CF on file pairs}\\
\hline
$F_{\mathrm{2T}}$ & 0.61 & 1.27 & 1.35\\
$F_{\mathrm{2TR}}$ & 0.61 & 1.27 & 1.26\\
 $F_{\mathrm{2STR}}$ & 0.69 & 1.27 & 1.23\\
\hline
\end{tabular}
\end{table}

Increasing the complexity of the transformation class did not affect the average compression factor over the individual melodies and {\em decreased\/} the compression factor over the pair files. This may be because more complex contrapuntal transformations, such as inversion, retrograde, augmentation and diminution, do not occur frequently enough in these melodies to compensate for the increase in transformation class complexity, $K(F)$, when going from $F_{\mathrm{2T}}$ to $F_{\mathrm{2TR}}$ to $F_{\mathrm{2STR}}$. $\fun{EncodePointSet}$ did, however, achieve a higher classification success rate with $F_{\mathrm{2STR}}$ than with the simpler transformation classes.

\section{Conclusions}
\label{conclusion-section}

We have presented an algorithm for discovering the maximal patterns in a dataset related by bijections within a user-specified transformation class. 
We have also described an algorithm that computes all the occurrences of these maximal patterns and uses these occurrence sets to construct a compressed encoding of the dataset. 
We have evaluated this compression algorithm with three different transformation classes on the task of classifying folk-song melodies into tune families. We found that, while the most complex transformation class gives the best classification success rate, it does not, on average, produce more compressed encodings on this corpus. This may be because the melodies in this corpus do not extensively employ more complex transformations like augmentation, diminution, inversion and retrograde. Despite this result, we believe the proposed approach 
is worthy of further investigation, since, in contrast to other current machine-learning approaches (e.g., neural network models), this approach computes {\em explanations\/} for data, in the sense that the encodings describe ways in which the data might have arisen as the result of transformations being applied to patterns within the data. The next step in our research will be to work on more thoroughly parallellising the proposed algorithms in order to be able to apply them to larger datasets in a variety of domains and on a broad range of machine-learning and information retrieval tasks. 

\bibliographystyle{splncs04}
\bibliography{bibliography}
\end{document}